%% file: paper.tex
\documentclass[]{ailab}

\usepackage{amsmath,amssymb}
\usepackage{graphicx}
\graphicspath{{figures/}{.}}
\usepackage{booktabs}
\usepackage{longtable}
\usepackage{natbib}
\usepackage{xspace}
\usepackage{float} 
\setcitestyle{square,comma,numbers,sort&compress}

\providecommand{\ours}{\textsc{Recursive Self-Rewrite}\xspace}

\providecommand{\oursabbr}{\textsc{RSR}\xspace}

\providecommand{\oursharness}{Recursive Self-Reflect Terminus\xspace}
\providecommand{\oursharnessabbr}{RSRT\xspace}

\title{\centering Scaling Trajectories for Complex Tasks through Recursive Self-Rewrite}

\author[1,2]{Zongxia Li\textsuperscript{$\dagger$}}
\author[1]{Yucheng Shi\textsuperscript{$\dagger$}}
\author[1,3]{Zhongzhi Li\textsuperscript{$\dagger$}}
\author[1,4]{Junyao Yang}
\author[1,5]{Ruhan Wang}
\author[6]{Chengsong Huang}
\author[2]{Fuxiao Liu}
\author[1]{Haitao Mi}
\author[7]{Jordan Boyd-Graber}
\author[1]{LeoweiLiang}

\affiliation[1]{Tencent HY LLM Frontier}
\affiliation[2]{University of Maryland, College Park}
\affiliation[3]{University of Georgia}
\affiliation[4]{National University of Singapore}
\affiliation[5]{Indiana University\protect\linebreak}
\affiliation[6]{Washington University in St.\ Louis}
\affiliation[7]{Nanyang Technological University}

\email{zli12321@terpmail.umd.edu, zhongzhili@global.tencent.com}

\contribution{$^\dagger$ Core contribution.}

\abstract{
Successful trajectories on difficult tasks are valuable supervision for model improvement.
Different harnesses enable the same model to solve these tasks in different ways.
Their successful trajectories provide useful experience for self-improvement, but they also contain controller interventions and workflow conventions that may be unavailable under a general harness.
We propose \ours{} (\oursabbr{}), a framework that converts these experiences into reusable model capabilities through recursive trajectory self-rewrite.
We use a single base model, Qwen-3.8-27B to discover successful solutions under diverse harnesses and rewrite them for learning under a general harness.
\oursabbr{} contains a \texttt{planner} that extracts useful procedures into runbooks, a \texttt{critic} that screens for verifier and solution leakage, and rejects candidates and recursively regenerate using critique feedback, and an \texttt{executor} that follows the qualified runbooks to solve each task in a fresh sandbox under a general harness.
We show that using three harnesses expands the range of task domains that Qwen-3.8-27B can solve and yields more valuable successful trajectories; with \oursabbr{}, we further reconstruct these into a larger set of high-quality trajectories.
We collect experience from approximately 3K self-curated terminal tasks. 
Across 3K tasks, the union of the three harnesses solves 759 tasks, which is 34.3\% more than the strongest individual harness in the recorded pool.
We further use \oursabbr{} to rewrite successful source trajectories, expanding the training set from 2,001 to 11,094 high-quality trajectories for finetuning Qwen-3.8-27B.
Training it on these rewritten trajectories outperforms both the base model and direct trajectory SFT: pass@3 increases from 57.0\% to 74.2\% on Terminal-Bench~2, from 1.5\% to 9.1\% on Terminal-Bench~4, from 39.0\% to 63.0\% on our self-curated Terminal-Bench Hard, and from 3.0\% to 6.0\% on our Software Terminal-Bench. On Long-Horizon Terminal-Bench, process reward rises from 0.21 to 0.29.
\par\smallskip
{\centering
\href{https://huggingface.co/IntelligenceLab/RSR-27B}{\raisebox{-0.2em}{\includegraphics[height=1.2em]{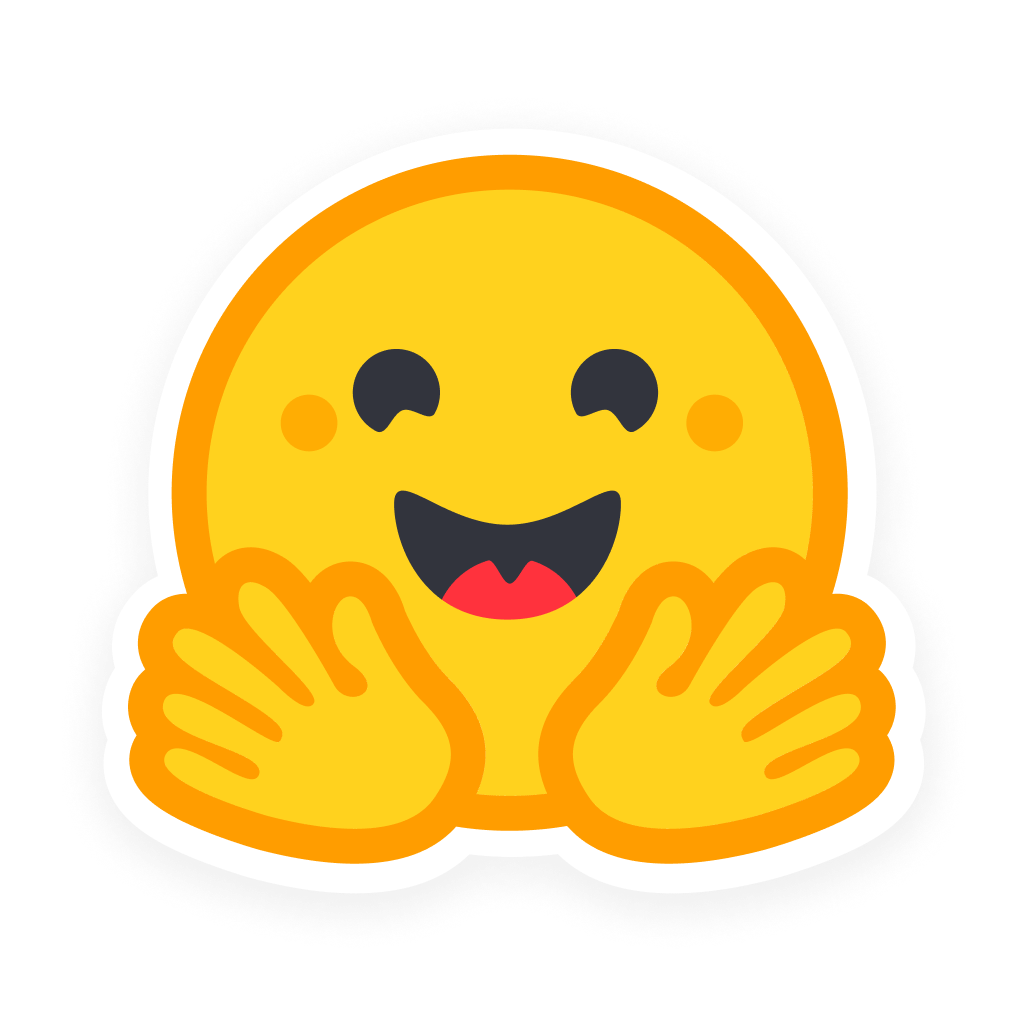}}\,\nolinkurl{https://huggingface.co/IntelligenceLab/RSR-27B}}\par}
}

\headercontent{\input{figures/concept_teaser}}

\date{\today}

\begin{document}
\thispagestyle{firstheader}
\maketitle

\input{sections/introduction}
\input{sections/method}
\input{sections/experiments}

\input{sections/related_work}
\input{sections/conclusion}
\bibliographystyle{plainnat}
\bibliography{refs}

\newpage
\appendix
\input{sections/appendix}

\end{document}

%% file: figures/concept_teaser.tex
\begin{minipage}{\linewidth}
  \color{metafg}
  \setlength{\parskip}{0pt}
  \captionsetup{type=figure,skip=5pt}
  \centering
  \includegraphics[width=\linewidth]{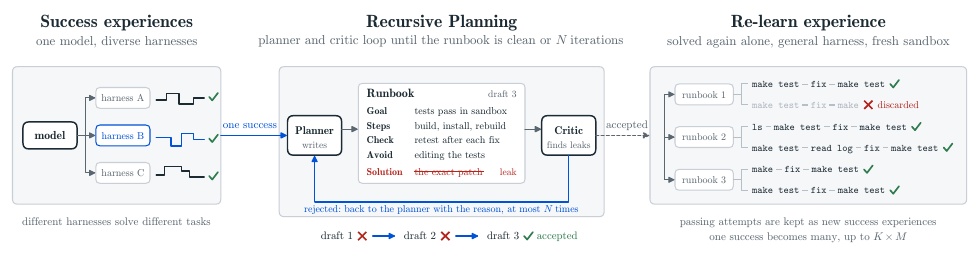}
  \caption*{\textbf{Recursive planning turns successful experience into new practice.}
  Diverse harnesses discover source successes. Planner--critic feedback refines runbooks within a retry budget; each screened runbook guides fresh executions under a general harness. Rewritten trajectories are verified and filtered before training. Commands and draft numbers are schematic examples.}
\end{minipage}

%% file: sections/introduction.tex
\section{Introduction}
\label{sec:introduction}

Human progress advances by turning experience into knowledge that others can reuse and by discovering new solutions.
Researchers approach problems with different tools, methods, and prior knowledge, producing successes, failures, and practical lessons. 
These lessons are organized into papers and textbooks so that later generations can learn from them without repeating the entire discovery process. 
We argue that AI agents can undergo a similar process through self-improvement, turning successful experience into reusable capability.

A model's capabilities depend not only on the model itself, but also on the harness through which it acts.
A harness determines how the model receives observations, invokes tools, tracks progress, verifies intermediate results, recovers from failure, and decides when to stop~\citep{yao2026harness}.
A growing body of work shows that improved harnesses can substantially increase terminal-task performance without updating model parameters~\citep{terminuskira2026,lee2026meta,lin2026agentic}. In some settings, stronger harnesses even enable weaker models to match or surpass stronger models, as demonstrated on terminal benchmarks and game-playing tasks~\citep{li2026statemem,lou2026autoharness}.
Different harnesses encourage different behaviors and are often effective for different task domains~\citep{yao2026harness,vats2026scaffoldeffect,evermind2026raven}.
For example, coding-oriented harnesses such as Codex~\citep{openai2025codex} and Claude Code~\citep{anthropic2025claudecode} are designed for coding tasks, goal-driven loops promote long-horizon exploration through repeated continuation~\citep{anthropic2025longrunning}, and StateM structures execution around explicit states and validated transitions~\citep{li2026statemem}.
As a result, different harnesses can solve complementary subsets of tasks, and access to multiple harnesses expands the set of tasks a model can solve (Section~\ref{sec:harness-coverage}).
When a model solves additional tasks under different harnesses, the resulting successful trajectories reveal behaviors and strategies that would not emerge under a single general-purpose harness, which these trajectories provide supervision for improving the model itself.

However, naively training on trajectories collected from different harnesses can introduce substantial distribution mismatch. 
Each trajectory reflects the joint behavior of the model and its source harness, including controller interventions, task-specific prompts, workflow logic, and stopping rules. Simply pooling these trajectories can mix harness-specific conventions rather than isolate the underlying problem-solving behaviors that made the solutions successful~\citep{thangarajah2026dcas}. 
As a result, a model trained on such data may rely on external control patterns that will be unavailable under a general harness at inference time. 
Effective self-improvement requires preserving useful planning, verification, recovery, and continuation behaviors while rewriting them into trajectories that are compatible with the general harness the model will ultimately use.


We propose \ours{}, a framework for self-improvement through multi-harness trajectory rewriting into a general harness, as shown in the opening illustration.
\ours{} first uses distinct harnesses to collect more diverse successful solutions across complementary parts of the task distribution. 
It then rewrites each solution using the same base model in three roles: a \textbf{planner} reconstructs the solution as a runbook that captures the key steps, checks, and recovery strategies: a \textbf{critic} removes verifier leakage and solution leakage, and an \textbf{executor} then follows the accepted runbook to solve the task again in a fresh sandbox under the general Terminus~2 harness~\citep{harborTerminus2}. 
We generate multiple runbooks and executions for each source solution and retain only passing trajectories through rejection sampling.
The entire pipeline uses a single base model, Qwen-3.8-27B~\cite{qwen38} end-to-end.

We use an internal RST pipeline~\cite{li2026recursive} to automatically curate $\sim$3K difficult terminal tasks, which spans 50+ domains including e.g, Science, Software, Hardware, Data. 
We use Terminus~2, StateM~\cite{li2026statemem}, and \oursharness{} (\oursharnessabbr{}) as discovery harnesses to collect approximately 2K successful trajectories covering 759 unique tasks.
These harnesses solve complementary subsets of tasks and broadens the range of successful experiences available for solving tasks. 
We then apply \ours{} with multiple executions per passing trajectory and generate approximately $\sim$10K verified trajectories for supervised finetuning.
Our results show that training on the rewritten trajectories improves Qwen-3.8-27B over Direct SFT by 20.8, 4.1, 4.6, 7.0, and 3.0 percentage points in pass@3 on Terminal-Bench~2 (TB2), Terminal-Bench~3 (TB3), Terminal-Bench~4 (TB4), Terminal-Bench Hard (TBH), and our Software Terminal Bench 100 (SWR~100), respectively. Mean per-run pass rates also improve on all five benchmarks, while process reward on Long-Horizon Terminal Bench (LHTB) increases from 0.25 to 0.29.
\ours{} show that different harnesses are not only useful as inference-time controllers, but also as tools for discovering experiences through which a model can improve its own general-purpose policy.

%% file: sections/method.tex
\section{\ours{}: Self-Improvement Through Trajectory Rewriting}
\label{sec:method}

\ours{} improves a model under a general-purpose harness by learning from successful trajectories discovered under diverse specialized harnesses. The key idea is that different harnesses help the same base model solve different difficult tasks, and those successful solutions can be rewritten into demonstrations that are compatible with a single general harness. 
We use the same base model throughout the process in three stages. 
First, we run the base model under multiple discovery harnesses and collect successful trajectories. 
Second, we rewrite each successful trajectory into new demonstrations under the general harness using three model roles: a \textbf{planner}, a \textbf{critic}, and an \textbf{executor}. 
Then we keep only verified rewritten trajectories as the training sources for model self-improvement. 
Figure~\ref{fig:trajectory-rewriting-pipeline} illustrates the overall process.
\input{figures/trajectory_rewriting_pipeline_figure}

\subsection{Multi-Harness Discovery}
\label{sec:method-discovery}
We use multiple discovery harnesses that differ in how they provide control and support during execution, such as progress tracking, continuation, validation, state management, or recovery from failure.
Because different harnesses can be effective on different types of tasks, they expand the range of successful solutions where the model itself with just one harness can solve.
%

\subsection{Trajectory Rewriting for Experince Learning}

Successful trajectories collected under different harnesses record how the model solves tasks with different workflow logic. 
Our goal is to turn these successful solutions into learning experiences that the model can practice and learn from under a general harness.
Intuitively, discovery helps the model solve a task once, while rewriting helps the model solve it again in a cleaner and more reusable way.
Trajectory rewriting involves three roles: a \texttt{planner} reconstructs the successful procedure as a runbook, a \texttt{critic} removes leakage and harness-specific information, and an \texttt{executor} follows the guidance to solve the task again under the general harness.
The resulting rewritten trajectories provide cleaner demonstrations through which the model can relearn the solution and improve its own policy.

\paragraph{Planner: Reconstructing Successful Solutions into Plans.}
The planner reconstructs a \emph{runbook} for each successful source trajectory: a structured description of how the task was solved. 
Before planning, we compact the source trajectory by keeping the task instruction, the model's actions, and the environment's observations, while removing harness-specific control messages.
The runbook summarizes the required end state, key milestones, useful checks, recovery strategies, and common pitfalls. 
To reduce direct answer transfer, runbooks may describe the task, relevant interfaces, and validation procedures, but they should not directly provide the finished deliverable. We sample multiple runbook candidates for each source trajectory.

\paragraph{Critic: Filtering Leakage.}
We use the mode itself as a critic to filter candidate runbooks before they are used for execution.
We first apply deterministic checks, such as schema validation, removal of known artifacts, and rejection of unsupported tool references.
We then use a model-based critic that sees only the public task instruction and the candidate runbook. Its role is to determine whether the runbook provides useful procedure or leaks information that the executor should not receive.
Only runbooks that pass this screening stage are kept for rewriting.\footnote{This part can be done with Jev for faster inference~\cite{deng2026jev}}

\paragraph{Executor: Relearning Successful Behaviors Under the General Harness.}
For each runbook, the executor re-solves the task in a fresh sandbox under the general harness. 
The runbook is provided as private guidance during generation, but it is never written into the public trajectory. 
The executor must therefore produce a new trajectory based on the current environment rather than replaying the source trajectory.



\paragraph{Trajectory Filtering.} 
We use the model itself to flag values that appear in a trajectory but cannot be derived from the task or the environment, a sign of hidden answer transfer, and discard the demonstrations that contain them.
For finetuning, we keep only the public interaction history (the task instruction, environment observations, and the model's responses) and remove the runbook and critic conversation.
At inference time, the finetuned model runs under the general-purpose harness alone, without the source harnesses or private runbooks, so any planning, checking, recovery, or continuation must come from the model itself.
\ours{} thus turns diverse harness-assisted successes into reusable experience for a single general-purpose harness.


%% file: figures/trajectory_rewriting_pipeline_figure.tex
\begin{figure}[t]
  \centering
  \includegraphics[width=\linewidth]{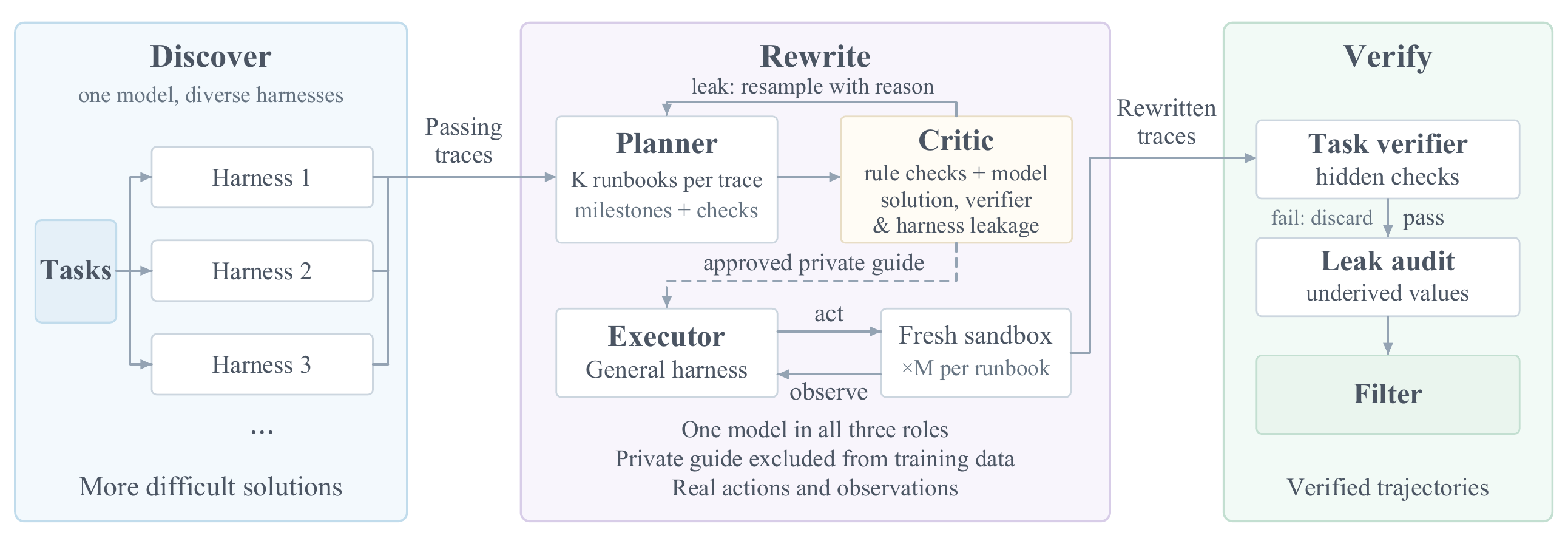}
  \caption{\textbf{Turning multi-harness successes into learning experiences with \ours{}.}
Different discovery harnesses enable the same model to solve more difficult tasks. \ours{} rewrites these harness-assisted solutions into demonstrations under a general harness, allowing the model to learn from successful experiences that a single harness alone would not produce. 
  }
  \label{fig:trajectory-rewriting-pipeline}
\end{figure}

%% file: sections/experiments.tex
\section{Experiments}
\label{sec:experiments}

We organize the experiments around the stages of self-improvement: collecting experience through diverse harnesses, reconstructing and auditing that experience, and training and evaluating the resulting model.

\subsection{Terminal Tasks}
\label{sec:exp-data}
We construct our training task pool from two sources. 
\textbf{SWR} is our self-constructed collection of approximately 2.5K terminal tasks spanning across {50} domains. 
These domains include software usage, biology, chemistry, physics, hardware, operations, and security. 
This breadth provides diverse settings in which the model must select tools, reason about environment feedback, and carry out task-specific procedures. 
We additionally include 420 filted and modified tasks from the \textbf{RST}~\cite{li2026recursive} dataset with increased difficulty. 
These sources define the approximately 3K-task pool. 
%
%

\subsection{Discovery Harnesses}
\label{sec:exp-harnesses}

\begin{figure}[t]
  \centering
  \includegraphics[width=\linewidth]{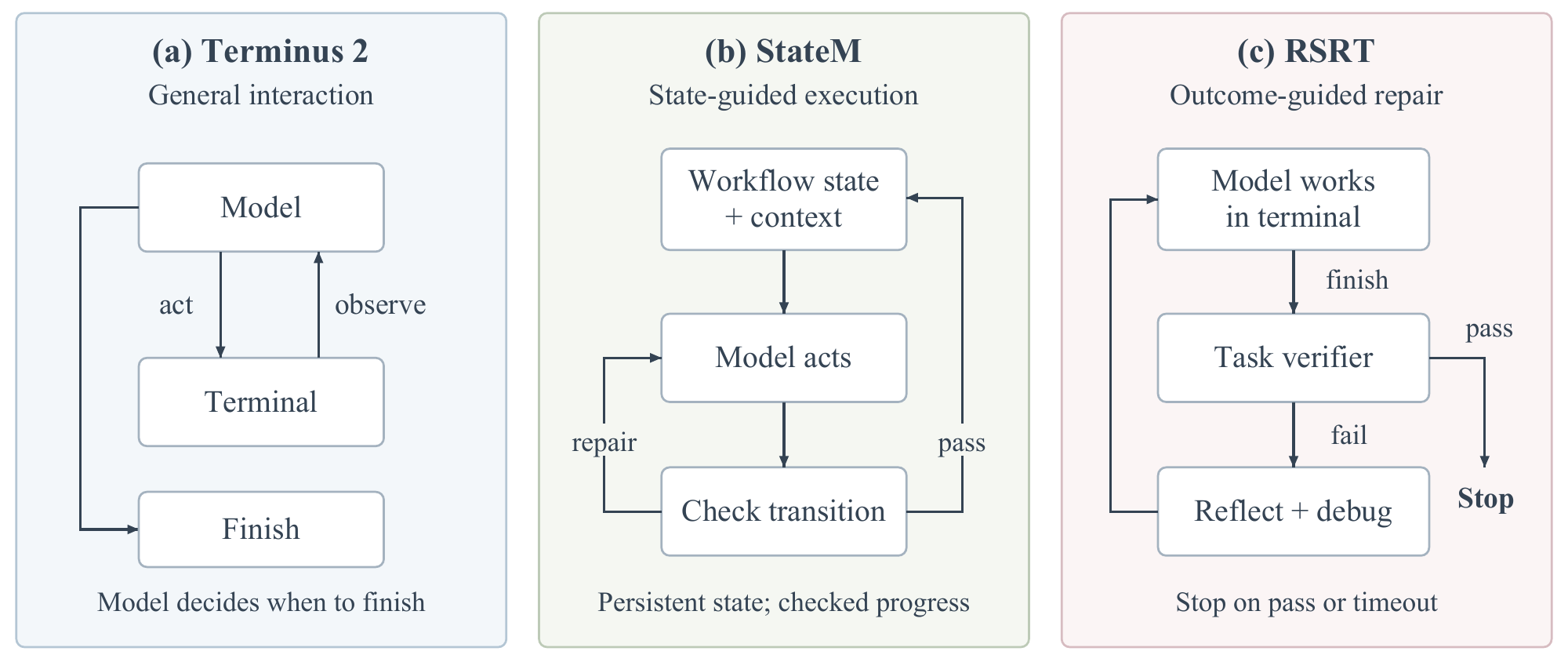}
  \caption{\textbf{Three harnesses represent different logics of solving problems.}
  Terminus~2 supports a general model--terminal interaction loop. StateM adds persistent workflow state and checked transitions. \oursharness{} uses verifier outcomes to trigger reflection and continued execution, stopping on success or timeout.}
  \label{fig:discovery-harnesses}
\end{figure}

We use three harnesses with different approaches to terminal problem solving: Terminus~2~\cite{harborTerminus2}, StateM~\cite{li2026statemem}, and our modified Terminus~2--\oursharness{} (\oursharnessabbr{}).
The underlying model is held fixed across harnesses, so the diversity of collected experience comes from differences in external execution support (Figure~\ref{fig:discovery-harnesses}). 

\paragraph{Terminus 2: general terminal interaction.}
Terminus~2 provides our general-purpose baseline harness~\citep{harborTerminus2}. The model issues terminal commands, observes their outputs, and decides how to proceed or when to finish. It provides a common interaction interface for collecting trajectories with relatively little specialized workflow control, and is also the target harness used for reconstructed executions.
This is also the general harness we use for model inference.

\paragraph{StateM: explicit workflow control.}
StateM organizes execution around persistent workflow states, phase-local context, and checked transitions~\citep{li2026statemem}. Recoverable runbooks and procedural practices help preserve progress and guide continuation after interruptions or failures. This structure supports problem solving in which the model must maintain execution state, satisfy intermediate conditions, and verify progress before moving to the next phase.

\paragraph{\oursharness{} (\oursharnessabbr{})}
\oursharness{} extends Terminus~2 to continue problem solving after unsuccessful completion attempts without revealing verifier internals or reference solutions, using only the final verification outcome as weak supervision.
The discovery protocol allows a wall-clock budget of up to 180 minutes per task.
When the model declares completion, the harness runs the task verifier on the current result. A passing result ends the rollout. 
If verification fails and time remains, the harness reports only the failure outcome and prompts the model to review its previous trajectory, identify possible mistakes, debug its work, and continue.
This process repeats until the task passes or the time limit is reached. The design encourages reflection and recovery beyond the model's initial stopping decision while keeping the verifier and the solution hidden.
Within a discovery rollout, continuation preserves the conversation and sandbox state. Trajectory reconstruction instead starts a new execution in a fresh sandbox under the general harness.

\input{sections/harness_coverage}
\input{sections/rewrite_analysis}

\subsection{Model Training and Evaluation}
\label{sec:exp-groups}
We now test the usefulness of the reconstructed experience trajectories on training it into the model.
We compare three groups based on the same Qwen-3.8-27B model. 
\textbf{Base} is the original model.
\textbf{Direct SFT} fine-tunes the model directly on the passing source rollouts collected with the discovery harnesses after removing harness-specific insertion phrases and parts, without runbook reconstruction or fresh re-execution.
\textbf{\ours{}} fine-tunes the model on verified rewrite trajectories, in addition to the 766 direct passing trajectories from Terminus~2 without rewriting. 

\textbf{Evaluation benchmarks.}
We evaluate on Terminal-Bench~2 (TB2)~\cite{merrill2026terminalbench}, Terminal-Bench~3 (TB3), Terminal-Bench~4 (TB4~\cite{marten2026terminalbench4}, Terminal-Bench Hard (TBH)~\cite{li2026recursive}, Long-Horizon Terminal Bench (LHTB)~\cite{li2026long}, and Software Terminal 100 (SWR~100).\footnote{Software Terminal is a self-curated benchmark.}

\subsection{Results and Discussion}
\label{sec:exp-main-results}
\label{sec:exp-discussion}
Table~\ref{tab:main-results} reports both pass@3 and the mean per-run pass rate over three runs for TB2, TB3, TB4, TBH, and SWR100. Pass@3 measures success within three attempts, while the mean averages the three individual-run pass rates. LHTB is evaluated using process reward. Parenthesized pass@3 counts are estimated from the reported percentages and benchmark sizes by rounding to the nearest integer; LHTB counts are reported task completions.
\oursabbr{} achieves higher pass@3 and mean per-run pass rate on all five benchmarks, as well as higher LHTB process reward, than Direct SFT.

\input{tables/main_results}

\paragraph{Diverse harness experiences still support model improvement.}
In pass@3, Direct SFT improves over Base by 17.0 percentage points on TBH, 5.4 points on TB~3, and 3.0 points on TB~4. It matches Base on SWR100, but falls by 3.6 points on TB~2. Its mean per-run pass rate on TB~2 also falls, from 51.7\% to 43.8\%, a decrease of 7.9 percentage points.
These mixed results suggest that diverse source experiences can improve performance even without rewriting.
However, direct SFT can also introduce undesirable behavior shifts. 
In particular, \oursharness{} produces many trajectories longer than 100 steps, which may be difficult for a 27B model to absorb without the original sustained-execution support.
In our inspection of three trajectories from the direct-SFT checkpoint, we observed a recurring \texttt{dead end} behavior: the model loops around the same action pattern without making real progress, which ultimately causes task failure.

\paragraph{Rewriting improves experience scaling.}
By scaling successful experiences and trajectories, and rewriting them under a standardized harness, \oursabbr{} further improves over Direct SFT by 20.8, 4.1, 4.6, 7.0, and 3.0 percentage points in pass@3 on TB~2, TB~3, TB~4, TBH, and SWR100, respectively.
The mean per-run results also favor \oursabbr{}: gains over Direct SFT are 26.3, 5.4, 2.9, 11.5, and 2.0 percentage points on the same five benchmarks, respectively. On TB~2, \oursabbr{} reaches 74.2\% pass@3 and 70.1\% mean per-run success; on TBH, it reaches 63.0\% and 54.0\%, respectively.
These results suggest that expanding experience and standardizing it under a general harness can produce better training samples for generalization.
The benefit also extends to process reward.
On LHTB, process reward increases from 0.21 for Base and 0.25 for Direct SFT to 0.29 for \oursabbr{}. All three groups still complete zero of the 46 tasks, so this gain reflects greater partial progress rather than more fully solved tasks.





\input{sections/harness_benchmark_comparison}

%% file: sections/harness_coverage.tex
\subsection{Multi-Harness Discovery and Rollouts}
\label{sec:harness-coverage}
\label{sec:exp-rollouts}


We use \textbf{Qwen-3.8-27B} to generate rollouts using three harnesses \textbf{Terminus~2}, \textbf{\oursharness{}}, and \textbf{StateM}. Then analyze the rollout domain coverage rates and pass rates by different harnesses.
We pool passing trajectories across the three harnesses as candidates for runbook reconstruction and fresh execution under the general harness. This collection stage provides experiences generated through different approaches to the same training problems.\footnote{Due to harness complexities of harness other than the general base harness, we have more sandbox errors rolling out from \oursharness{} and StateM, leading to fewer rollouts on \oursharness{} than other harnesses.}



We generated a total of 14,598 rollouts from the dataset, including 2,001 successful trajectories covering 759 distinct tasks.
We distinguish \emph{task coverage}, the fraction of tasks solved by at least one recorded rollout, from \emph{rollout success rate}, the fraction of individual attempts that pass. 
Multiple passing trajectories can correspond to the same task. 

\paragraph{Using different harnesses expands the model's successful solution coverage.}
Table~\ref{tab:harness-coverage} shows the tasks solved by each harness and by their unions. 
%
The three-harness union solves a total of 759 tasks (25.9\%), adding 194 solved tasks with a relative increase of 34.3\%. The union improves over the strongest individual harness by 19.4\% on RST and 33.8\% on SWR. 
These additional solved tasks broaden the experiences available and available successful trajectories for trajectory reconstruction.

\begin{table*}[t]
  \centering
  \small
  \setlength{\tabcolsep}{5pt}
  \begin{tabular}{p{0.29\textwidth}rrrrrr}
    \toprule
    & \multicolumn{2}{c}{RST (420)} &
      \multicolumn{2}{c}{SWR (2,509)} &
      \multicolumn{2}{c}{Pooled (2,929)} \\
    \cmidrule(lr){2-3}\cmidrule(lr){4-5}\cmidrule(lr){6-7}
    Harness & Solved & Unique & Solved & Unique & Solved & Unique \\
    \midrule
    Terminus~2 & 94 (22.4\%) & 8 & 471 (18.8\%) & 121 & 565 (19.3\%) & 129 \\
    \oursharnessabbr{} & 108 (25.7\%) & 14 & 404 (16.1\%) & 77 & 512 (17.5\%) & 91 \\
    StateM & 95 (22.6\%) & 10 & 366 (14.6\%) & 58 & 461 (15.7\%) & 68 \\
    \midrule
    \shortstack[l]{Terminus~2 $\cup$\\\oursharnessabbr{}} & 119 (28.3\%) & 34 & 572 (22.8\%) & 264 & 691 (23.6\%) & 298 \\
    Terminus~2 $\cup$ StateM & 115 (27.4\%) & 21 & 553 (22.0\%) & 226 & 668 (22.8\%) & 247 \\
    \oursharnessabbr{} $\cup$ StateM & 121 (28.8\%) & 35 & 509 (20.3\%) & 159 & 630 (21.5\%) & 194 \\
    \midrule
    \textbf{All three} & \textbf{129 (30.7\%)} & \textbf{+21} & \textbf{630 (25.1\%)} & \textbf{+159} & \textbf{759 (25.9\%)} & \textbf{+194} \\
    \bottomrule
  \end{tabular}
  \caption{\textbf{The union of three harnesses lead to the most unique tasks solved.}
  \emph{Unique} denotes tasks solved by only that harness. For the final row, ``+'' denotes the additional tasks solved relative to the strongest individual harness. A task is counted as solved when at least one recorded rollout passes. Rollout counts differ across harnesses in the full pool due to sandbox errors or task errors.}
  \label{tab:harness-coverage}
\end{table*}

\begin{figure*}[t]
  \centering
  \includegraphics[width=\textwidth]{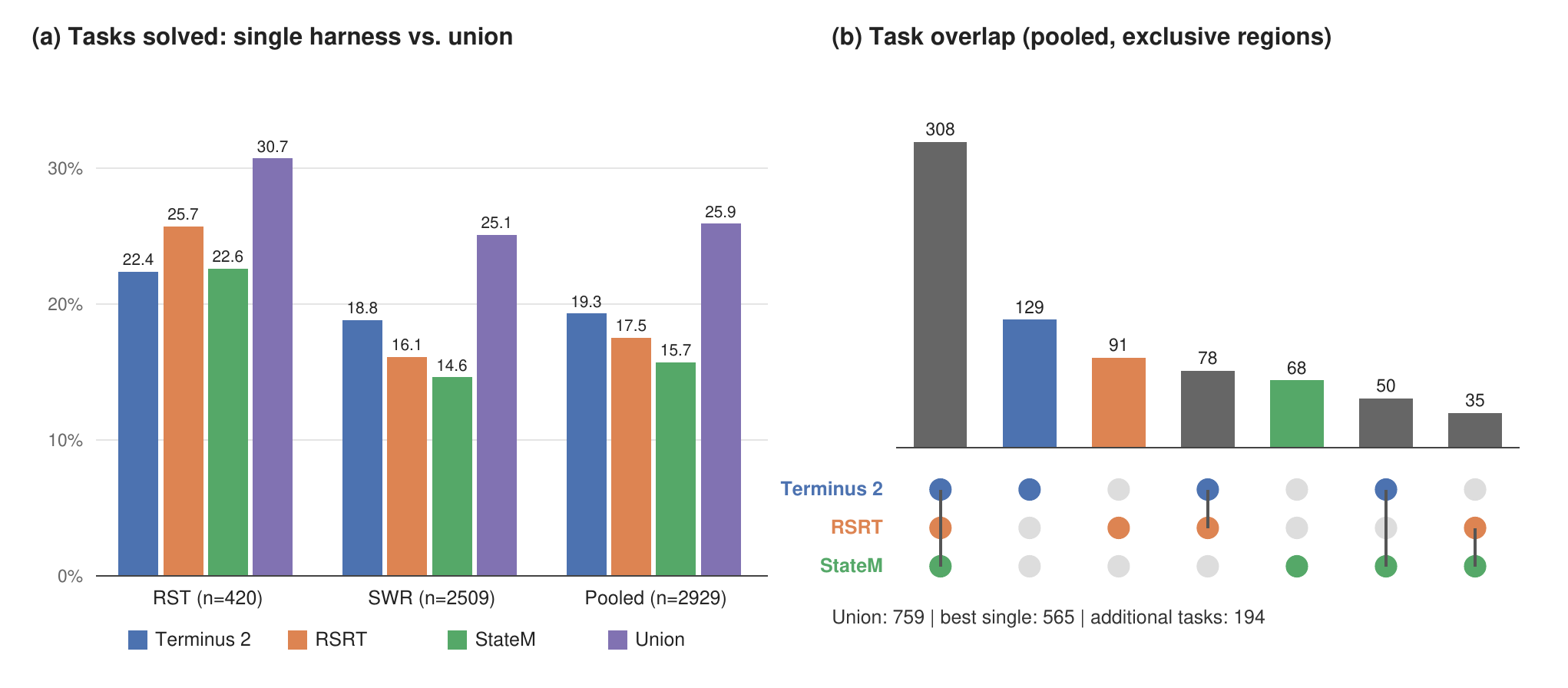}
  \caption{\textbf{Complementary task coverage across harnesses.}
  (a) Task coverage of each harness and their union on our terminal tasks, and the pooled task set.
  (b) Exclusive and overlapping solve regions on the pooled set. Colored dots indicate the harnesses included in each region. Of the 2,929 tasks, 2,170 are not solved by any harness. Coverage uses all recorded rollouts; per-harness rollout counts differ.}
  \label{fig:harness-coverage-union}
\end{figure*}

Among the 759 tasks solved by at least one harness, 288 are solved exclusively by one harness: 129 only by Terminus~2, 91 only by \oursharness{}, and 68 only by StateM. Another 163 are solved by exactly two harnesses, while 308 are solved by all three. Every pairwise union exceeds the coverage of the strongest individual harness, and each third harness contributes further solved tasks (Figure~\ref{fig:harness-coverage-union}).


\begin{table}[t]
  \centering
  \small
  \setlength{\tabcolsep}{6pt}
  \begin{tabular}{lrrrr}
    \toprule
    Harness & Rollouts & Passing & Success rate & Tasks solved \\
    \midrule
    Terminus~2 & 5,405 & 766 & 14.2\% & 565 \\
    \oursharnessabbr{} & 3,777 & 599 & 15.9\% & 512 \\
    StateM & 5,416 & 636 & 11.7\% & 461 \\
    \midrule
    Pooled & 14,598 & 2,001 & 13.7\% & 759 \\
    \bottomrule
  \end{tabular}
  \caption{\textbf{Discovery rollouts on the terminal task set.}
  Passing counts successful trajectories. Tasks solved counts distinct tasks;
  the pooled row uses the union across harnesses. Success rate is per rollout.}
  \label{tab:discovery-rollouts}
\end{table}

\paragraph{Complementarity persists with equal rollout counts.}
The full pool contains 5,405 Terminus~2, 3,777 \oursharness{}, and 5,416 StateM rollouts, so its individual-harness ranking is affected by unequal sampling (Table~\ref{tab:discovery-rollouts}).
We therefore also examine the 1,247 tasks with equal rollout counts across harnesses: 420 RST tasks with one rollout per harness and 827 SWR tasks with two, totaling 2,074 rollouts per harness. 
On this subset, Terminus~2, \oursharness{}, and StateM solve 249 (20.0\%), 285 (22.9\%), and 227 (18.2\%) tasks, respectively. Their union solves 352 (28.2\%), adding 67 tasks over \oursharness{}, a relative increase of 23.5\%.
Each harness still contributes exclusive solves: 30, 56, and 21. 

\paragraph{Different harnesses complement different domain problems.}
We analyze the doamin and task-type coverage from the three harnesses.
The dataset contains 87 fine-grained labels: 35 RST task categories and 52 SWR tool families. 
These labels differ from the broad scientific and engineering domains. 
Under this taxonomy, Terminus~2, \oursharnessabbr{}, and StateM solve at least one task in 49, 46, and 42 domains, respectively, whereas their union reaches 51.
Across 88 task types, they cover 56, 52, and 49 types individually, while their union covers 60 domains.

\begin{figure*}[t]
  \centering
  \includegraphics[width=0.92\textwidth]{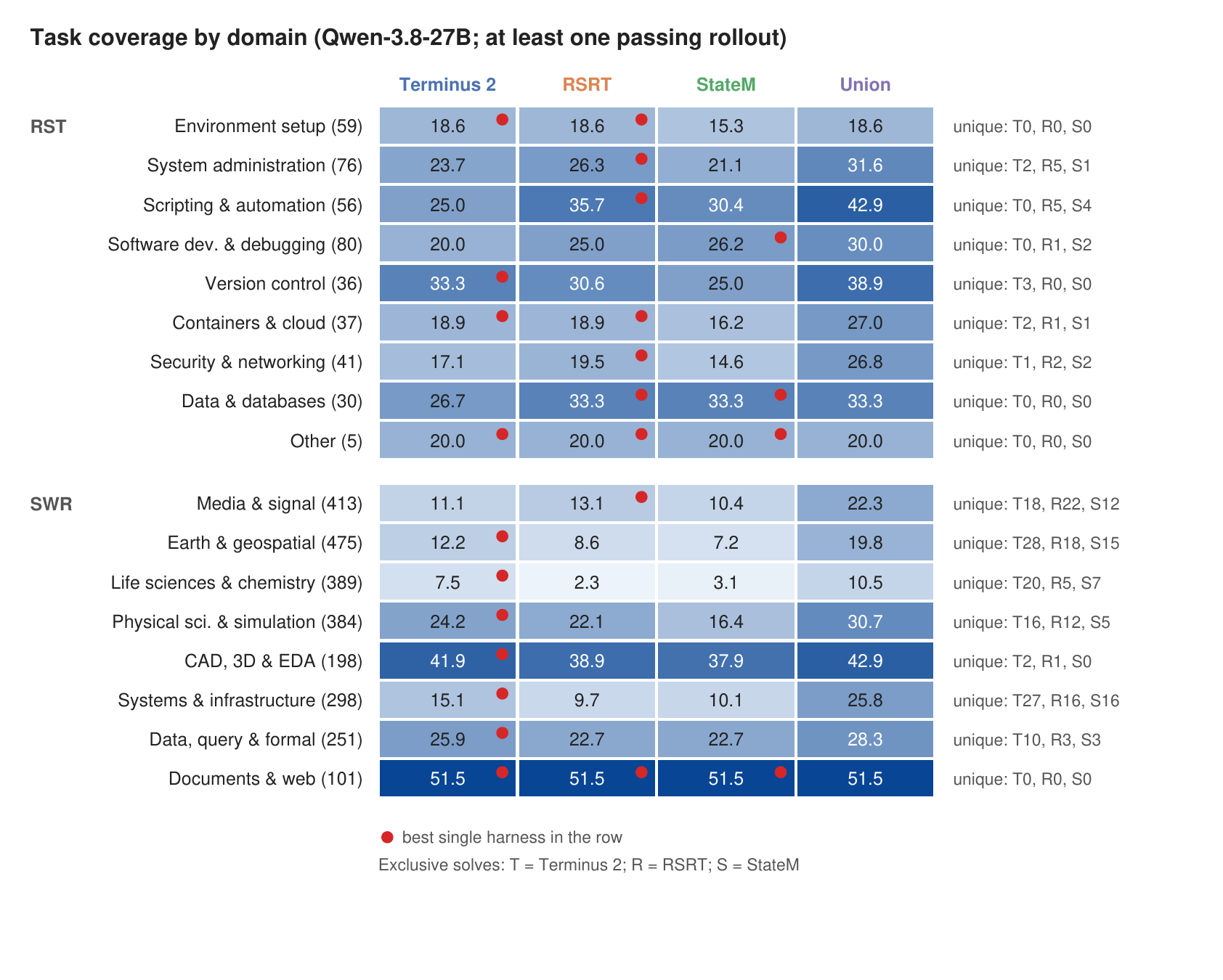}
  \caption{\textbf{Different harness covers differnt domains of expertise.}
  Cell values report the percentage of tasks with at least one passing rollout, using all recorded rollouts on tasks evaluated by all three harnesses. A red marker identifies the strongest individual harness in each row. The final column reports the union, and the right-hand annotations report tasks solved exclusively by Terminus~2 (T), \oursharnessabbr{} (R), or StateM (S).}
  \label{fig:harness-domain-coverage}
\end{figure*}

Figure~\ref{fig:harness-domain-coverage} shows that no individual harness has the highest observed coverage in every domain. 
\oursharnessabbr{} leads on RST scripting and automation (35.7\%) and SWR media and signal processing (13.1\%). StateM leads on RST software development and debugging (26.2\%), while Terminus~2 leads on RST version control and several SWR scientific domains.
Combining harnesses raises SWR media and signal coverage to 22.3\% and earth and geospatial coverage to 19.8\%, compared with best individual rates of 13.1\% and 12.2\%. 
%

\paragraph{Successful discovery trajectories differ in execution style.}
The 2,001 passing trajectories also exhibit different execution styles (Table~\ref{tab:harness-execution-style}). 
Terminus~2 has a median of 17 turns and assigns 55.5\% of commands to exploration. 
\oursharness{} has a median of 18 turns and averages 2.20 completion claims per trajectory; 4.5\% of its passing trajectories contain a rejected completion before eventual success. StateM has a median of 24 turns and devotes 29.6\% of commands to state-management operations.
In the fixed sample of 200 failed trajectories per harness (100 from each data source), \oursharness{} averages 67.1 turns, compared with 22.8 for Terminus~2 and 23.4 for StateM.
Thus, extended execution can recover some tasks while also spending substantial effort on unsuccessful attempts. 


\begin{table}[t]
  \centering
  \small
  \setlength{\tabcolsep}{4pt}
  \begin{tabular}{lrrr}
    \toprule
    & Terminus~2 & \oursharnessabbr{} & StateM \\
    \midrule
    Passing trajectories & 766 & 599 & 636 \\
    Turns (mean) & 19.4 & 23.2 & 24.0 \\
    Turns (median) & 17.0 & 18.0 & 24.0 \\
    Commands per turn & 2.10 & 1.96 & 2.37 \\
    Completion claims per trajectory & 1.72 & 2.20 & 1.11 \\
    Passes after rejection (\%) & 0.0 & 4.5 & 0.0 \\
    \midrule
    Exploration (\%) & 55.5 & 50.6 & 35.3 \\
    Python execution (\%) & 21.8 & 22.1 & 12.6 \\
    State management (\%) & 0.0 & 0.0 & 29.6 \\
    File writing (\%) & 9.0 & 8.6 & 5.7 \\
    Testing and verification (\%) & 5.0 & 5.1 & 4.0 \\
    \bottomrule
  \end{tabular}
  \caption{\textbf{Execution styles on passing trajectories.}
  All passing rollouts on the common task set are included. Command categories are assigned by the audit's rule-based classifier and reported as percentages of issued commands.}
  \label{tab:harness-execution-style}
\end{table}

\paragraph{Task-level comparisons illustrate possible sources of complementarity.}
\oursharnessabbr{} uniquely solves a system\hyp{}administration task by resuming after a rejected completion, removing residual processes, and correcting the required log. 
StateM uniquely solves a multi-requirement software task by enumerating requirements in a contract and catching an exact dependency-version constraint missed by the other harnesses. 
Terminus~2 uniquely solves a watershed reconstruction task through direct exploration and implementation. 
These task-level examples illustrate how continued execution, structured checks, and direct exploration can address different failure modes. 


%% file: sections/rewrite_analysis.tex
\subsection{Recursive Self-Rewrite and Case Study}
\label{sec:rewrite-analysis}

Using diverse harnesses broadens the successful experiences available to a single model.
We then rewrite these experiences for learning under the general Terminus~2 harness using Qwen-3.8-27B in the planner, critic, and executor roles.
Specifically, we reconstruct successful rollouts from StateM and \oursharness{} so that their useful procedures are expressed through the general Terminus~2 interface.
We sample $K=4$ candidate runbooks, screen them with the critic, and use each retained runbook to guide $M=4$ fresh executions under the general harness at temperature $0.7$ for each successful source rollout,.
We keep only passing rewrites that satisfy the filtering requirements in Section~\ref{sec:method}, remove private runbooks and critic conversations, and use the remaining public execution histories as training trajectories.
We collect a total of 12,893 rollouts with \ours{} covering 975 unique tasks, with passing rewrites of a total $86\%$, a total of available passing 11,094 rewrites.

\input{sections/case_studies}

%% file: sections/case_studies.tex
\begin{figure}[!htb]
  \centering
  \includegraphics[width=\linewidth]{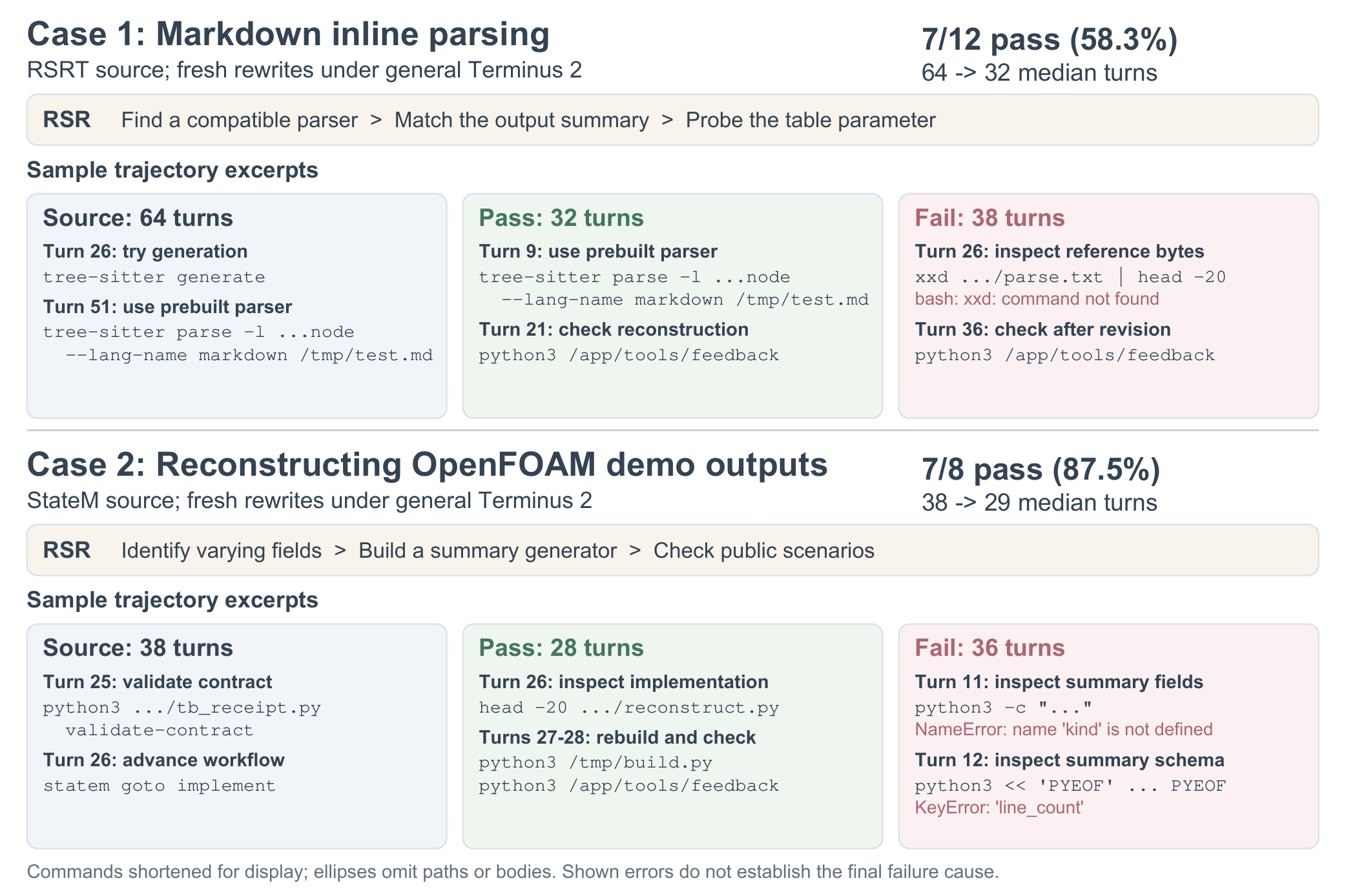}
  \caption{\textbf{From source experience to successful and unsuccessful rewrites.}
  Each case compares a source trajectory with a passing and a failing rewrite guided by the same runbook. Headers report success rates and median passing lengths across all recorded rewrites (12 Markdown; eight OpenFOAM). Commands are abbreviated; displayed errors do not establish the cause of final verifier failure. Verifier success does not imply quality-audit acceptance.}
  \label{fig:rewrite-case-studies}
\end{figure}

We examine two rewritten tasks to show how a source experience can be rewritten into new trajectories under the general Terminus~2 harness: Markdown inline parsing from \oursharness{} and OpenFOAM PitzDaily reconstruction from StateM. 
Figure~\ref{fig:rewrite-case-studies} shows a passing and a failing execution guided by the same runbook in each case, with representative command excerpts from the source and both rewrites. 
These are independent sampled executions. 
We count parsed assistant turns, including command-free turns, consistently across sources and rewrites.\footnote{The parsed sources contain 64 and 38 turns, respectively; their registry entries report 65 and 39. We use the parsed histories for both the traces and length comparisons.}

\paragraph{Markdown task: earlier setup does not guarantee successful reconstruction.}
For the Markdown inline-parsing task, the source trajectory spends many turns on setup, package compatibility, repeatedly calling the same tool across multiple turns, and parser selection before reaching a workable solution. 
Rewritten executions avoid much of this early exploration by directly reusing guidance about the compatible parser and key reconstruction steps. Even so, success is not guaranteed: some rewrites pass while others fail after similarly short executions. Across 12 rewrites, 7 pass (58.3\%). The median passing rewrite is 32 turns, a 50.0\% reduction from the 64-turn source. 
However, this case shows that shortening the setup phase does not remove the core reconstruction difficulty.

\paragraph{OpenFOAM: fresh execution retains analysis and repair.}
For the OpenFOAM PitzDaily task, the source trajectory interleaves analysis, implementation, and structured checking. 
Rewritten executions under the general harness preserve this pattern, which they continue to analyze outputs, revise implementations, and repair errors during execution, but they do so without issuing source-harness-specific commands. 
Across 8 rewrites, 7 pass (87.5\%). The median passing rewrite is 29 turns, although some successful rewrites are still longer than the source. 
This case shows that rewriting can preserve useful analysis and repair behavior while shifting execution to the general interface.

These cases show that runbooks can transfer useful procedures into fresh executions under a general harness. 
They also show that the same guidance can still lead to different outcomes, so shorter or cleaner executions do not by themselves guarantee success. 

\input{sections/case_command_similarity}

%% file: sections/case_command_similarity.tex
\subsubsection{Command Similarity Across Rewrites}
\label{sec:case-command-similarity}

We compare pairs of rewrites guided by the \emph{same runbook} with pairs guided by \emph{different runbooks} for each task in Figure~\ref{fig:rewrite-case-studies}, including both passes and failures. 
Markdown has 12 rewrites from three runbooks, yielding 18 same-runbook and 48 different-runbook pairs. 
OpenFOAM has eight rewrites from two runbooks, yielding 12 and 16 pairs. 
We average each metric separately within these groups.

\paragraph{Metrics.}
We use several trajectory-structure metrics to compare executions. 
Exact command overlap is the Jaccard similarity between sets of recorded command fingerprints, where shell whitespace is normalized but script bodies remain distinct. 
Command-order similarity is defined as $2L/(n_i+n_j)$, where $L$ is the longest common subsequence of command fingerprints and $n_i,n_j$ are the command counts. Tool-set overlap compares the sets of invoked program names. Tool-transition overlap applies Jaccard similarity to distinct ordered pairs of adjacent commands' tool sets. Action-sequence similarity is defined as $1-d/\max(n_i,n_j)$, where $d$ is the unit-cost Levenshtein distance between sequences of heuristic action labels such as exploration, Python execution, and verification. Commands are ordered across turns. All scores are scaled to 0--100 and measure execution structure rather than semantic equivalence.

\input{tables/case_command_similarity}

\paragraph{Shared runbooks increase command overlap.}
For Markdown, trajectories from the same runbook show higher exact command overlap (19.1\% vs.\ 12.1\%) and command-order similarity (24.6\% vs.\ 15.7\%) than trajectories from different runbooks (Table~\ref{tab:case-command-similarity}). OpenFOAM shows the same pattern, though more weakly: 12.6\% vs.\ 10.4\% for exact overlap and 18.2\% vs.\ 14.7\% for command order. After excluding the first two turns, within-runbook exact overlap remains higher in both cases.

\paragraph{Workflow similarity is more task-dependent.}
Markdown also shows higher within-runbook tool-transition overlap (34.0\% vs.\ 18.9\%) and action-sequence similarity (53.6\% vs.\ 42.0\%). OpenFOAM shows little difference in tool transitions (28.4\% vs.\ 28.5\%), and action-sequence similarity is slightly lower within runbooks (61.0\% vs.\ 63.1\%). Thus, shared runbooks are associated with more similar command-level behavior, while broader workflow similarity depends more on the task.

%% file: tables/case_command_similarity.tex
\begin{table}[H]
  \centering
  \small
  \setlength{\tabcolsep}{5pt}
  \renewcommand{\arraystretch}{1.15}
  \begin{tabular}{lrrrr}
    \toprule
    & \multicolumn{2}{c}{Markdown (12 rewrites)} & \multicolumn{2}{c}{OpenFOAM (8 rewrites)} \\
    \cmidrule(lr){2-3}\cmidrule(lr){4-5}
    Similarity metric & Same Runbook & Different Runbook & Same Runbook & Different Runbook \\
    \midrule
    Pairs compared& 18 & 48 & 12 & 16 \\
    \midrule
    Exact command overlap (Jaccard) & 19.1 & 12.1 & 12.6 & 10.4 \\
    Command order (normalized LCS) & 24.6 & 15.7 & 18.2 & 14.7 \\
    Tool-set overlap (Jaccard) & 66.9 & 52.9 & 83.3 & 84.4 \\
    Tool-transition overlap (Jaccard) & 34.0 & 18.9 & 28.4 & 28.5 \\
    Action-sequence similarity (edit) & 53.6 & 42.0 & 61.0 & 63.1 \\
    \midrule
    Exact overlap, omit first two turns & 12.5 & 6.2 & 7.2 & 4.7 \\
    \bottomrule
  \end{tabular}
  \caption{\textbf{Command similarity within and across runbooks.}
  Same pairs share a runbook; different pairs use distinct runbooks for the same task.
  Scores are mean pairwise similarities on a 0--100 scale, including both passing
  and failing rewrites. Higher means more similar. The final row excludes each
  rewrite's first two assistant turns.}
  \label{tab:case-command-similarity}
\end{table}

%% file: tables/main_results.tex
\begin{table}[!htbp]
  \centering
  \caption{\textbf{\oursabbr{} achieves the best results across benchmarks.}
  We report pass@3 and the mean per-run pass rate over three runs for TB2, TB3, TB4, TBH, and SWR100, and process reward for LHTB. In addition, Direct SFT performs worse on TB 2 after training due to circular and deadend behaviors in raw trajectories from \oursabbr{}.}
  \label{tab:main-results}
  \small
  \setlength{\tabcolsep}{7pt}
  \renewcommand{\arraystretch}{1.2}
  \begin{tabular*}{\linewidth}{@{\extracolsep{\fill}}llccc@{}}
    \toprule
    Benchmark & Metric & Base & Direct SFT & \oursabbr{} \\
    \midrule
    TB 2 & Pass@3
      & 57.0\% (51/89)
      & 53.4\% (48/89)
      & \textbf{74.2\%} (66/89) \\
    TB 3 & Pass@3
      & 0.0\% (0/74)
      & 5.4\% (4/74)
      & \textbf{9.5\%} (7/74) \\
    TB 4 & Pass@3
      & 1.5\% (1/66)
      & 4.5\% (3/66)
      & \textbf{9.1\%} (6/66) \\
    TBH (ours) & Pass@3
      & 39.0\% (39/100)
      & 56.0\% (56/100)
      & \textbf{63\%} (63/100) \\
    SWR100 (ours) & Pass@3
      & 3.0\% (3/100)
      & 3.0\% (3/100)
      & \textbf{6.0\%} (6/100) \\
    \midrule
    TB 2 & Mean (3 runs)
      & 51.7\% & 43.8\% & \textbf{70.1\%} \\
    TB 3 & Mean (3 runs)
      & 0.0\% & 1.4\% & \textbf{6.8\%} \\
    TB 4 & Mean (3 runs)
      & 0.0\% & 3.2\% & \textbf{6.1\%} \\
    TBH (ours) & Mean (3 runs)
      & 31.0\% & 42.5\% & \textbf{54.0\%} \\
    SWR100 (ours) & Mean (3 runs)
      & 3.0\% & 3.0\% & \textbf{5.0\%} \\
    \midrule
    LHTB (ours) & Process reward
      & 0.21 (0/46)
      & 0.25 (0/46)
      & \textbf{0.29} (0/46) \\
    \bottomrule
  \end{tabular*}
\end{table}

%% file: sections/harness_benchmark_comparison.tex
\subsection{Ablation Studies of Harnesses on Sample Benchmarks}
\label{sec:harness-benchmark-comparison}

To further examine whether different harnesses expose different problem-solving strengths on benchmarks, we evaluate Qwen-3.8-27B base model under Terminus~2, StateM, and \oursharness{} on TB2, TBH, and TB3. 
This comparison examines the effect of changing the execution harness, complementing the training comparison in Section~\ref{sec:exp-main-results} on benchmarks without training.
Here, \oursharnessabbr{} denotes the discovery harness from Section~\ref{sec:exp-harnesses}, while \oursabbr{} denotes the rewriting method.
We run once on each benchmark using each harness and show results in Table~\ref{tab:harness-benchmark-comparison}.

\input{tables/harness_benchmark_comparison}

\paragraph{Smaller models can solve challenging tasks under stronger harnesses.}
For the Qwen-3.8-27B base model, TBH and TB3 remain challenging benchmarks, and the set of passing tasks varies substantially across harnesses.
This further suggests that different harnesses can unlock different problem-solving behaviors even for the same model.
StateM, by contrast, is better aligned with TB2 and achieves the highest pass@1 on that benchmark. This shows that the same base model can solve more challenging tasks when paired with a more suitable harness, even without additional training. 
\ours{} builds on this observation by collecting more such successful trajectories and using them to further improve the model through training.


%% file: tables/harness_benchmark_comparison.tex
\begin{table}[H]
  \centering
  \caption{\textbf{Different harnesses also affect Qwen-3.8-27B Base Model performance on different benchmarks.}
  All rows use the same base model without additional training. Entries show reported success rates (\%) with the supplied task counts in parentheses.}
  \label{tab:harness-benchmark-comparison}
  \small
  \setlength{\tabcolsep}{6pt}
  \renewcommand{\arraystretch}{1.15}
  \begin{tabular*}{\linewidth}{@{\extracolsep{\fill}}lrrr@{}}
    \toprule
    Harness & TB2 (89) & TBH (100) & TB3 (74) \\
    \midrule
    Terminus~2 & 51.7 (46) & 33.0 (33) & 0.0 (0) \\
    \oursharness{} & 55.1 (49) & \textbf{66.0} (66) & \textbf{4.1} (3) \\
    StateM & \textbf{58.4} (52) & 46.0 (46) & 1.3 (1) \\
    \midrule
    Union of all three
      & \textbf{68.5} (61)
      & \textbf{70.0} (70)
      & \textbf{4.3} (3) \\
    \bottomrule
  \end{tabular*}
\end{table}

%% file: sections/related_work.tex
\section{Related Work}
\label{sec:related}
\paragraph{Harness engineering and optimization.}
A growing line of work improves agent performance by changing the execution harness while keeping model weights fixed. 
Harnesses structure planning, tool use, progress tracking, verification, and recovery, helping models complete goals and sustain long-horizon tasks that they struggle to solve under less structured execution~\citep{li2026statemem,wang2026rethinking,merrill2026terminalbench}.
Terminal-Bench highlights this dependence by evaluating executable command-line tasks, where success reflects both the model and its interaction system~\citep{merrill2026terminalbench}. 
Terminus-KIRA, StateM, Deep Agents, Ledger, Meta-Harness, Agentic Harness Engineering, Self-Harness, and AutoSaddler improve this system through stronger verification, continuation, workflow control, or automated harness search~\citep{terminuskira2026,li2026statemem,trivedy2026deepagents,merrill2026terminalbench,lee2026metaharness,lin2026agentic,zhang2026selfharness,park2026autosaddler}.
Harness changes avoid model retraining, they offer a relatively inexpensive route to higher solve rates and can make weaker models more competitive. 
Better harnesses can even reverse model rankings on particular benchmarks, enabling a weaker or lower-cost model to outperform a stronger model operating without the same support~\citep{lou2026autoharness,li2026statemem}.
Self-Harness improves fixed models on Terminal-Bench by changing only the harness, raising MiniMax M2.5 from 40.5\% to 61.9\% and Qwen3.5-35B-A3B from 23.8\% to 38.1\% on its held-out split~\citep{zhang2026selfharness}. 
StateM similarly reports large gains on Terminal-Bench~2.1, including improvements for GPT-5.6 Luna and DeepSeek-V4 Flash~\citep{li2026statemem}.
Beyond terminal tasks, AutoHarness and AI4AI at Test-Time show that stronger external control can substantially improve weaker models without parameter updates~\citep{lou2026autoharness,qian2026ai4ai}. 
Task-CoEvolve further reduces the cost of harness optimization through adaptive validation-task selection~\citep{miyai2026taskcoevolve}. 
Controlled studies such as the Scaffold Effect and Harness-Bench show that harness choice affects both solve rate and efficiency, motivating its inclusion in model-performance reporting~\citep{vats2026scaffoldeffect,yao2026harness}. 
Our complementary objective is not to optimize a single harness, but to use those diverse harnesses as discovery tools and maximize a weak models' ability to solve more tasks.


\paragraph{Recursive self-improvement and Self-Evolving of agent systems.}
Recursive self-improvement work uses past execution results to iteratively improve an agent's model, harness, or both~\cite{zheng2026dreamrsi,li2026mmzeroselfevolvingmultimodelvision,ICLR2026_d49b9aac,he2025visplayselfevolvingvisionlanguagemodels}.
ModularRSI~\cite{wu2026modularrsi} and SoL-Pi~\cite{liu2026solpi} analyze trajectories to revise harness modules or improve execution efficiency, while Dream-RSI optimizes exploration-policy code through replay of recorded discovery trees~\citep{zheng2026dreamrsi}.
ScienceBuddy couples these ideas more tightly by adapting the harness in an inner loop and training the model in an outer loop~\citep{xue2026sciencebuddy}. 
These methods improve the execution system itself, or co-evolve the model and harness together. 
In contrast, \ours{} does not optimize a single harness or assume an autonomous recursive cycle. 
Instead, it collects successful solutions from multiple harnesses and reconstructs them as verified demonstrations under a shared general interface, so that the model can improve through supervised learning from its own harness-assisted experience.

\paragraph{Learning from trajectories and harness-supported behaviors.}
Another related line of work studies how execution experience can be reused as contextual guidance or as direct training signal.
Reflexion and ExpeL retain reflective lessons from earlier attempts and retrieve them during later problem solving without updating model weights~\citep{shinn2023reflexion,zhao2023expel}.
SWE-Gym provides executable software-engineering tasks and verified successful trajectories for policy learning~\citep{pan2024swegym}.
Other work studies how behaviors initially supported by external guidance can become part of the model itself, including SKILL0, PATS, and EvoHarness-RL~\citep{lu2026skill0,shi2026pats,ning2026evoharness}.
Cross-harness training work further shows that trajectory compatibility matters: DCAS studies transfer across scaffolds, while \citet{yu2026coevolving} show that directly imitating stronger experts can misfit a weaker model's execution setting~\citep{thangarajah2026dcas}.
Harness-Zero most directly studies transferring harness-supported behaviors into model parameters for deployment under a fixed target harness~\citep{ye2026harness}.
\ours{} aims to preserve useful behaviors after specialized support is removed. 
Our focus is on reconstructing completed successes from diverse harnesses into verified demonstrations under a shared general harness with model's own ability.

%% file: sections/conclusion.tex
\section{Conclusion}
\label{sec:conclusion}

\ours{} improves a model through its own experiences collected under diverse problem-solving harnesses while generating more valuable problem-solving trajectories.
Different harnesses help the same model discover more complementary successful solutions, but their trajectories often contain harness-specific execution conventions that do not transfer directly to a general harness.
\oursabbr{} addresses this gap by reconstructing those successful experiences into new verified trajectories under a general harness.
Our analysis and experiments show that diverse harnesses expand the set of tasks a model can solve, and that rewriting these experiences helps the model relearn useful planning, checking, recovery, and sustained-execution behaviors under a general harness.
These results suggest that diverse harnesses are not only effective tools for discovery, but also valuable sources of experience through which models can improve themselves on complex long-horizon tasks that are difficult to solve or annotate at scale.

%% file: sections/appendix.tex
\input{sections/appendix_task_list}